\documentclass[letterpaper, 10 pt, journal, twoside]{IEEEtran}
\ifCLASSINFOpdf
\else
\fi
\usepackage{graphics} % for pdf, bitmapped graphics files
\usepackage{epsfig} % for postscript graphics files
\usepackage{multicol}
\usepackage{multirow}
\usepackage{amsmath} % assumes amsmath package installed
\usepackage{amssymb}  % assumes amsmath package installed
\usepackage{hyperref}

\begin{document}
\title{EASE: Embodied Active Event Perception via Self-Supervised Energy Minimization}

% \author{Michael~Shell,~\IEEEmembership{Member,~IEEE,}
%         John~Doe,~\IEEEmembership{Fellow,~OSA,}
%         and~Jane~Doe,~\IEEEmembership{Life~Fellow,~IEEE}% <-this % stops a space
% \thanks{M. Shell was with the Department
% of Electrical and Computer Engineering, Georgia Institute of Technology, Atlanta,
% GA, 30332 USA e-mail: (see http://www.michaelshell.org/contact.html).}% <-this % stops a space
% \thanks{J. Doe and J. Doe are with Anonymous University.}% <-this % stops a space
% \thanks{Manuscript received April 19, 2005; revised August 26, 2015.}}
\author{Zhou Chen$^{1}$, Sanjoy Kundu$^{1}$, Harsimran S. Baweja$^{2}$ and Sathyanarayanan N. Aakur$^{1}$%
\thanks{Manuscript received: January, 12, 2025; Revised March, 23, 2025; Accepted June, 11, 2025.}%Use only for final RAL version
\thanks{This paper was recommended for publication by Editor Tetsuya Ogata upon evaluation of the Associate Editor and Reviewers' comments.
This work was partially supported by the US National Science Foundation Grants IIS 2348689 and IIS 2348690 and the US Department of Agriculture grant 2023-69014-39716-1030191.)} %Use only for final RAL version
\thanks{$^{1}$Z. Chen, S. Kundu, and SN Aakur are with the CSSE Department, Auburn University, Auburn, Alabama, USA, 36849. $^{2}$HS Baweja is with the School of Kinesiology, Auburn University, Auburn, Alabama, USA, 36849. \\
        E-mails: {\tt\footnotesize \{zzc0053,szk0266,hsb0025,san0028\}@auburn.edu}%
}
}

\markboth{IEEE Robotics and Automation Letters. Preprint Version. June, 2025}
{Chen \MakeLowercase{\textit{et al.}}: EASE: Active Event Perception} 




\maketitle

\begin{abstract}
Active event perception, the ability to dynamically detect, track, and summarize events in real time, is essential for embodied intelligence in tasks such as human-AI collaboration, assistive robotics, and autonomous navigation. However, existing approaches often depend on predefined action spaces, annotated datasets, and extrinsic rewards, limiting their adaptability and scalability in dynamic, real-world scenarios. Inspired by cognitive theories of event perception and predictive coding, we propose EASE, a self-supervised framework that unifies spatiotemporal representation learning and embodied control through free energy minimization. EASE leverages prediction errors and entropy as intrinsic signals to segment events, summarize observations, and actively track salient actors, operating without explicit annotations or external rewards. By coupling a generative perception model with an action-driven control policy, EASE dynamically aligns predictions with observations, enabling emergent behaviors such as implicit memory, target continuity, and adaptability to novel environments. Extensive evaluations in simulation and real-world settings demonstrate EASE's ability to achieve privacy-preserving and scalable event perception, providing a robust foundation for embodied systems in unscripted, dynamic tasks. 
\end{abstract}

% \begin{IEEEkeywords}
% IEEE, IEEEtran, journal, \LaTeX, paper, template.
% \end{IEEEkeywords}

% \fi
\IEEEpeerreviewmaketitle

\section{INTRODUCTION}
Understanding and responding to dynamic events is a cornerstone of embodied perception, yet it remains a significant challenge for autonomous agents. 
% Conventional event perception models rely on predefined action categories, extensive annotations, or task-specific priors, limiting adaptability to novel and unscripted scenarios. 
While recent work, such as Gumbsch et al.~\cite{gumbsch2023learning}, explores hierarchical temporal abstraction through discrete world models, event perception in real-world settings still predominantly relies on supervised or weakly supervised segmentation and tracking pipelines, limiting adaptability to novel, dynamic environments. 
Additionally, most tracking systems store raw video frames or extracted features for post-processing, raising privacy concerns in sensitive applications such as healthcare, surveillance, and assistive robotics. While encryption can secure stored data, it does not eliminate the risks associated with data retention. Instead, real-world applications demand agents that can perceive, track, and summarize events in real-time without storing personally identifiable information (PII) or relying on external supervision. 
Despite advances in vision and low-cost hardware, achieving robust event perception remains difficult due to the need for continuous adaptation in dynamic environments. Privacy constraints restrict reliance on stored data, and existing methods often treat event segmentation and motor control as separate tasks, ignoring their natural coupling. Addressing these challenges requires a unified framework that integrates perception and control, operates in real-time, and adapts to novel scenarios without compromising privacy. 

\begin{figure}[t]
    % \centering
    \includegraphics[width=0.95\columnwidth]{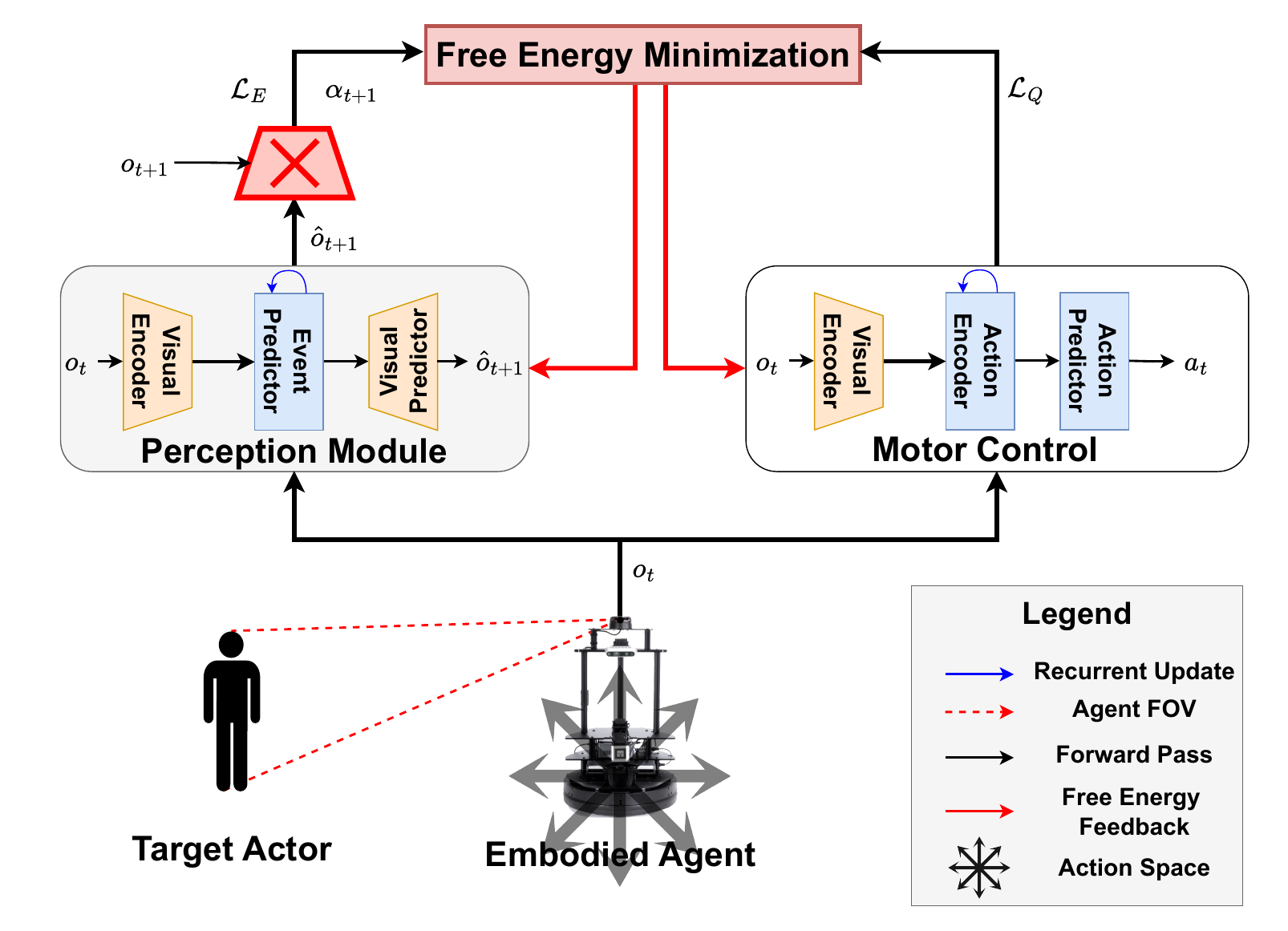}
    \caption{
    \textbf{Overview.} The perception module predicts future observations ($o_{t+1}$) and minimizes discrepancies ($\mathcal{L}_E$), while the motor control module selects actions ($a_t$), choosing from discrete motion primitives (forward, backward, stop, turn left, turn right, forward-left, and forward-right) to reduce control loss ($\mathcal{L}_Q$). {Free Energy Minimization} supervises both modules, enabling robust \textit{active} event perception. 
    }
    \label{fig:arch}
\end{figure}

In this work, we propose EASE, a novel framework for active event perception that unifies spatiotemporal representation learning and embodied control through a predictive learning-based interpretation of free energy minimization. Unlike classical active inference models that explicitly maintain probabilistic beliefs over latent states, EASE leverages predictive coding-inspired mechanisms to dynamically align perception and action by minimizing perceptual prediction errors. 
Drawing inspiration from cognitive theories of event perception \cite{zacks2001event} and visuomotor control \cite{idei2022emergence}, EASE learns to segment, summarize, and track events without requiring annotations or extrinsic rewards. The framework consists of two tightly integrated modules: a perception module that predicts future observations and quantifies uncertainty and a motor control module that selects actions to reduce this uncertainty by aligning predictions with observations. Together, these modules operationalize free energy minimization through a design that separates perception-driven prediction errors and action-driven uncertainty reduction, enabling EASE to iteratively refine event representations while adapting to dynamic environments. Prior work has primarily treated event perception \cite{Aakur2018APP,Aakur2020ActionLT,Trehan2024SelfsupervisedMS} and visuomotor control \cite{zhong2018advat,zhong2021advat,trehan2022towards} as separate problems, often relying on predefined action spaces, annotated datasets, or task-specific priors. In contrast, EASE offers a self-supervised, unified approach that enables embodied agents to autonomously learn event representations and optimize control policies in real-time, without predefined supervision.

Specifically, \textbf{our contributions} are as follows:
(i) we propose a unified framework for embodied active event perception that integrates generic event segmentation, summarization, and active tracking using prediction error and entropy as intrinsic signals,
(ii) we develop a free energy minimization paradigm that couples perception and action to enable dynamic adaptation to high-uncertainty regions and salient actors,
(iii) we introduce an inherently privacy-preserving snapshot summarization strategy that retains only salient events while discarding redundant or sensitive data, and
(iv) we validate EASE through extensive experiments in both simulation and real-world settings, demonstrating its effectiveness across tasks such as generic boundary detection, active tracking, and summarization without requiring annotations or external supervision.

\section{RELATED WORK}
\textbf{Active tracking} has been extensively explored in prior literature through approaches focused on \textit{task-specific supervision} or \textit{multi-agent systems}, differing from our self-supervised, unified event perception framework. 
Wei \emph{et al.}\cite{Wei2019AnIM} improved UAV-based target searching using a modified DQN with depth and segmentation priors. Trehan \emph{et al.}\cite{trehan2022towards} introduced an energy-based framework combining predictive learning with PID control for active action localization. Batista \emph{et al.}\cite{Batista2021TrainingAA} proposed a transfer learning framework for tracking in simulation, while Luo \emph{et al.}\cite{Luo2017EndtoendAO} developed a CNN-LSTM model linking frame-level inputs to camera control, differing from EASE’s predictive error-driven behaviors. 
Other works address distractors and collaborative tracking. Zhong \emph{et al.}\cite{Zhong2021TowardsDA} introduced a distraction-robust tracker using a multi-agent game with cross-modal learning. Li \emph{et al.}\cite{Li2020PoseAssistedMC} and Fang \emph{et al.}\cite{Fang2022CoordinateAlignedMC} extended tracking to multi-camera systems with pose-assisted collaboration and reinforcement learning. Adversarial training methods \cite{Zhong2018ADVATAA,zhong2021advat} employ tracker-target dueling to improve robustness via domain-specific rewards and multi-stage training.

\textbf{Event perception} has largely focused on offline event detection and action localization in pre-recorded videos, often requiring extensive training and lacking active control integration. Self-supervised methods~\cite{Aakur2020ActionLT,Trehan2024SelfsupervisedMS} localize action boundaries using predictive learning, while weakly supervised approaches~\cite{Mettes2018PointlySupervisedAL,Nguyen_2018_CVPR,Nguyen_2019_ICCV,Shi_2020_CVPR} bypass annotations through attention mechanisms and generative models. Recent works~\cite{Hu_2024_CVPR,Zhang2024CanMG} incorporate multi-modal and language models to enhance action localization in unstructured data. Although self-supervised segmentation frameworks~\cite{Aakur2018APP,Mounir2021SpatioTemporalES,aakur2020unsupervised} support taxonomy-free event detection, they focus on representation learning rather than embodied perception. Tracking models~\cite{cetintas2023unifying} complement traditional event perception by maintaining spatial continuity of dynamic entities, enabling more stable segmentation and summarization of evolving events. However, they typically assume passive observation and lack action-driven adaptation, making them unsuitable for active tracking in dynamic, interactive environments. 

\textbf{Self-supervised learning} (SSL) has become a key approach in computer vision, enabling robust representation learning without manual annotations. Common pretext tasks include image colorization~\cite{Larsson2017ColorizationAA}, solving jigsaw puzzles~\cite{kim2018learning}, and, for videos, future frame prediction~\cite{Aakur2018APP, Mounir2021SpatioTemporalES, Aakur2020ActionLT}, pace estimation~\cite{wang2020self}, or visual tempo understanding~\cite{yang2020video}. Recently, contrastive learning~\cite{dave2022tclr, kuang2021video} and auto-encoding approaches~\cite{sun2023masked} have dominated SSL, emphasizing distinctions between data samples or reconstruction of masked regions. 
The \textbf{free energy principle} (FEP), rooted in cognitive science, provides a unified framework for predictive coding and active inference~\cite{friston2010free, friston2012active,oliver2021empirical,lanillos2021active}, framing perception and action as processes that minimize prediction error. FEP-inspired methods have been applied to dynamic systems such as human-robot interaction~\cite{pezzulo2018active}, navigation~\cite{parr2022active}, and humanoid robot control~\cite{lanillos2021active}, offering biologically grounded solutions. Recent works~\cite{mazzaglia2022free,fujii2024real,matsumoto2023incremental} have further explored integrating FEP with deep learning, demonstrating its applicability to embodied agents. However, these approaches primarily adapt variational free energy formulations for latent state inference or task-driven active inference. 
In this paper, we introduce EASE, a self-supervised framework that unifies event perception and active control through free energy minimization, eliminating the need for annotations, predefined tasks, or adversarial rewards. 
Unlike classical FEP-based methods that rely on explicit probabilistic state estimation, our framework enables real-time tracking and segmentation by dynamically aligning perception with action-driven uncertainty reduction. This allows EASE to function in streaming, open-ended environments without requiring predefined state transitions or extensive probabilistic inference. 
% 
% Unlike task-specific tracking or multi-agent methods, EASE leverages intrinsic prediction errors for adaptive tracking and segmentation without supervision. Unlike offline event perception models, it operates in a streaming, embodied context, integrating segmentation, summarization, and tracking. 
% By directly leveraging self-supervised signals for real-time action and perception, EASE bridges biologically inspired principles with modern embodied systems.

% \clearpage
\section{EASE: EMBODIED ACTIVE SPATIOTEMPORAL EVENT PERCEPTION}

\textbf{Overview.} Our framework, EASE, consists of two subsystems that work together for sensory event perception and motor control. The overall architecture is illustrated in Figure~\ref{fig:arch}. A \textit{perception module} processes a sequence of sensory observations ($\{o_t\}_{t=1}^T$; $o_t \in \mathbb{R}^{H \times W \times C}$) to generate intrinsic signals for event perception and motor control in the form of spatiotemporal uncertainty distributions ($\alpha_t$) and temporal segmentation cues ($\delta_t$). These signals guide the \textit{motor control module}, which produces a sequence of actions ($\{a_t\}_{t=1}^T$; $a_t \in \mathbb{R}^k$) to minimize the system's energy and stabilize event representations. 
While traditional transition models~\cite{friston2010free, friston2012active} explicitly predict multi-step state evolution conditioned on actions, EASE employs a lightweight one-step prediction mechanism, where uncertainty indirectly influences future observations through action selection. 
The motor control module selects actions that direct the agent toward high-uncertainty regions for iterative event refinement without an explicit transition model. These subsystems operate in an \textit{active inference loop}, where perception informs action, and actions, in turn, refine sensory inputs to reshape event representations without external supervision, thereby learning robust control policies. 

\subsection{Learning as Free Energy Minimization}
Our framework integrates perception and action to minimize predictive uncertainty, drawing inspiration from active inference~\cite{friston2010free,friston2012active} and predictive coding~\cite{zacks2001event}. 
% Unlike classical free energy formulations, which explicitly model probability distributions over latent states and perform variational inference, EASE directly minimizes perceptual prediction errors to guide action selection. 
While classical free energy formulations involve variational inference over latent states with precision-weighted prediction errors, EASE adopts a simplified instantiation by directly minimizing squared L2 prediction error, corresponding to a unit-precision (identity covariance) assumption. Formally, this is defined as $\mathcal{L}_E = \|\mathbf{f}_{t+1} - \hat{\mathbf{f}}_{t+1}\|_2^2$. 
The perception module refines event representations by anticipating sensory changes and detecting salient features via uncertainty distributions. In contrast, the motor control module reduces uncertainty by selecting actions that direct the agent toward high-uncertainty regions. This enables real-time adaptation and interactive event segmentation without requiring an explicit generative model, predefined task structures, or handcrafted rewards.

We formulate this as an uncertainty-minimizing objective with two complementary terms: a \textit{prediction-based drive}, which quantifies the discrepancy between observed and predicted sensory inputs, and an \textit{action-driven uncertainty reduction} term, which captures how actions influence future prediction errors. Unlike conventional free energy approaches that treat variational free energy as a probabilistic bound on evidence, our formulation directly leverages uncertainty estimation to drive active control:

\begin{equation}
    \arg \min_a \left[ \|o(a) - \hat{o}\|^2 + \lambda \sum_{i,j} \alpha_{ij}(a) \|o_{ij}(a) - \hat{o}_{ij}\|^2 \right]
\end{equation}

% \noindent where $o(a)$ represents the sensory observation at time $t$ influenced by action $a$, $\hat{o}$ is the predicted observation from the perception module, and $\alpha_{ij}(a)$ represents the spatial uncertainty distribution at different spatial positions $(i,j)$, dynamically modulated by action $a$ to focus on high-uncertainty regions. 
\noindent where $o(a)$ represents the sensory observation at time $t$ influenced by action $a$, $\hat{o}$ is the predicted observation from the perception module, and $\alpha_{ij}(a)$ denotes the spatial uncertainty at location $(i,j)$ on the CNN feature map (e.g., $14 \times 14$ grid over the image). These values guide the agent to focus on high-uncertainty regions for targeted action modulation. 
The {first term} represents global perceptual prediction error, refining the system’s internal model of sensory dynamics. The second term encourages selecting actions that minimize uncertainty by reducing surprise in salient regions, where the uncertainty $\alpha_{ij}(a)$ is derived from the distribution of prediction errors to align the feature space with their corresponding spatial regions. 
The factor $\lambda$ is a design choice rather than a tunable hyperparameter, bridging perception and action. We fix $\lambda{=}1$ to fully integrate self-supervised perceptual feedback with motor control, ensuring actions directly influence future predictions. Setting $\lambda{=}0$ reduces the framework to a passive tracker without motor feedback. Intermediate values lack a clear interpretation, as the two terms operate at different timescales—prediction errors drive long-term event perception, while motor control optimizes short-term uncertainty reduction. 
% This distinction ensures that EASE maintains a coherent balance between perception-driven event modeling and action-driven adaptation.

\subsubsection{Prediction-based Drive} 
The perception module learns the spatiotemporal dynamics of the environment through a recurrent, generative model. Rather than operating directly at the frame level, the perception module processes input at the \textit{feature level}, leveraging a hierarchical predictor network to anticipate spatiotemporal patterns in a structured and computationally efficient manner. This approach enables the system to model complex temporal dependencies, capturing fine-grained spatial details and long-term event patterns critical for robust event learning and motor control. 
A visual encoder $\phi(o_t)\longrightarrow \mathbf{f}_t$ encodes the raw visual observation into a spatial feature map that captures spatial representations of the scene at time $t$. 
At each timestep, the perception module processes the feature map $\mathbf{f}_t \in \mathbb{R}^{h \times w \times d}$ and predicts the expected future feature map $\hat{\mathbf{f}}_{t+1}$. 
The anticipated feature map is compared to the actual features to compute the prediction error: $\mathcal{L}_{E} = \|\mathbf{f}_{t+1} - \hat{\mathbf{f}}_{t+1}\|^2$. This prediction error serves as the primary intrinsic signal driving the system, allowing it to refine its generative model continuously. 
% By operating at the feature level, the perception module abstracts away unnecessary details while focusing on high-level patterns, ensuring robustness across varying viewpoints and dynamic environmental changes caused by the agent's motion.

\textit{Quantifying uncertainty.} The prediction errors generated by the perception module also provide a mechanism for capturing uncertainty and guiding focus. Discrepancies between observed and predicted feature maps highlight areas where the system’s understanding is incomplete or inaccurate. These spatially distributed errors compute an uncertainty distribution $\alpha_{ij}$ that dynamically allocates focus to salient regions. The uncertainty distribution is computed by 
\begin{equation}
\alpha_{ij} = \text{Softmax}\left(\frac{\|\mathbf{f}_{t,ij} - \hat{\mathbf{f}}_{t,ij}\|^2}{\tau}\right), \text{where} f_t=\phi(o_t)    
\end{equation}

\noindent where $\mathbf{f}_{t,ij}$ and $\hat{\mathbf{f}}_{t,ij}$ represent feature vectors at spatial location $(i, j)$, and $\tau$ controls sensitivity to prediction errors. 
The softmax function normalizes the raw prediction errors into a spatial uncertainty distribution, ensuring that higher uncertainty regions receive greater emphasis while preserving relative differences across the feature map. 
This uncertainty quantification enables the system to dynamically track and follow regions of high prediction error, such as a moving person, by allocating focus to areas where the model's understanding is incomplete. 
By prioritizing these high-uncertainty regions, the system continuously refines its generative model to maintain accurate and context-aware event perception, allowing it to segment and interpret spatiotemporal dynamics associated with human activity. 

\subsubsection{Uncertainty-based Action Selection}
The motor control module leverages the uncertainty distribution \(\alpha_{ij}\) to guide action selection by learning a policy that aligns the center of the frame with regions of high prediction error (areas of high \(\alpha_{ij}\)). 
As in traditional active inference, actions are selected to minimize the system's uncertainty (surprise) by focusing on areas where the generative model struggles to make accurate predictions, implicitly minimizing the free energy over time. 
The motor control policy is parametrized by a neural network that shares the visual encoder of the perception module, allowing the motor control module to directly utilize the feature-level representation \(\mathbf{f}_t\). A Deep Q-Network (DQN) architecture~\cite{chung2013playing} is used to compute Q-values for a discrete set of actions \(a_t \in \mathcal{A}\). The reward function driving the learning process is defined as $r_t{=}-\|c_t{-}u_t\|$, where \(c_t\) is the center of the frame at time \(t\), and \(u_t\) is the location of the highest value in the uncertainty distribution \(\alpha_{ij}\). The term \(-\|c_t {-} u_t\|\) measures the proximity of the frame center to the region of highest uncertainty, encouraging the system to focus on high-\(\alpha_{ij}\) regions. Over time, this alignment reduces the system’s prediction error and refines the generative model. 
The DQN takes as input the state \(s_t {=} \{\mathbf{f}_t, \alpha_{ij}\}\), comprising the feature map \(\mathbf{f}_t\) and the uncertainty distribution \(\alpha_{ij}\), and outputs Q-values for each action. These Q-values represent the expected cumulative reward for selecting a given action. The policy is trained using a temporal difference loss:
\begin{equation}
\mathcal{L}_{Q} = \mathbb{E}\left[\left(Q(s_t, a_t) - \left(r_t + \gamma \max_{a'} Q(s_{t+1}, a')\right)\right)^2\right]
\end{equation}
where \(r_t\) is the reward and \(\gamma\) is the discount factor for future rewards.  
Unlike classical reinforcement learning approaches that rely on extrinsic task rewards, our framework uses intrinsic uncertainty to guide action selection, ensuring that the agent continuously improves its perceptual and motor capabilities. 
This reward-driven action selection mechanism dynamically reduces uncertainty while closing the perception-action loop for robust event perception and adaptive interaction.

\subsubsection{Learning Process}
The learning process jointly optimizes the perception and motor control modules to minimize the free energy defined in Equation (1). The prediction loss, \(\mathcal{L}_{E}\), reduces discrepancies between observed and predicted feature maps, addressing the first term of Equation (1). The motor control policy minimizes the temporal difference loss $
\mathcal{L}_{Q}$ aligns the framework's observations with the region of highest uncertainty. 
These losses are optimized jointly using shared feature representations, ensuring that perception and action are coupled to reduce global prediction error and guide the agent to high-uncertainty regions. Gradient-based updates refine the generative model and the motor control policy, dynamically reducing free energy and enabling robust event perception and adaptive motor control.

\subsubsection{Implementation Details}
We use the Stable-Baselines3~\cite
{raffin2021stable} DQN implementation with the following key parameters: a batch size of $32$, a replay buffer size of 50,000, a learning rate $10^{-5}$, and an exploration final epsilon of $0.02$. The policy network includes a custom feature extractor based on the first eight layers of EfficientNet-B0~\cite{tan2019efficientnet}, outputting feature maps of size $(320, 7, 7)$. These features are processed by 2 LSTM modules: LSTM-Event, which models temporal dynamics and outputs a (320, 7, 7) feature map, and LSTM-QVal, which aggregates temporal features into a 1024-dimensional vector for the policy. We add two fully connected layers with 256 and 64 units and an output layer to match the action space size.  
% Training updates the model iteratively based on a two-stage strategy. 
Early training ($t {<} 50,000$) focuses on aligning predicted and actual features, while later stages optimize temporal difference loss using computed intrinsic rewards. The framework is trained for $300k$ timesteps on UnrealCV-Gym~\cite{qiu2017unrealcv}.

\subsection{Event Perception: Segmentation and Summarization}
Generic event segmentation and snapshot creation are fundamental tasks for embodied agents navigating dynamic environments. These tasks empower agents to structure streaming visual input into meaningful temporal segments and summarize observations into concise, representative snapshots. Unlike conventional methods that rely on predefined event categories, our approach is inherently adaptive, capturing novel activities and ensuring scalability to unconstrained scenarios. Importantly, this framework also supports privacy-preserving behavior by focusing only on salient events and discarding redundant or sensitive information. Leveraging the free energy formulation, the system integrates perception and action to dynamically detect transitions in uncertainty and generate interpretable event summaries. 
% This enables the agent to adaptively perceive, process, and retain meaningful information while balancing interpretability, adaptability, and privacy concerns.

To enable generic event segmentation and summarization from streaming videos, the framework leverages prediction errors (\(\mathcal{L}_E\)) and entropy to detect event boundaries and select representative frames. This process is grounded in the free energy minimization framework, where segments are identified in regions of high uncertainty, and summarization minimizes local prediction errors within those segments. 
The system detects event boundaries \(B_t\) based on the entropy of prediction errors within a sliding window of size \(N\):
\begin{equation}
    B_t = \arg \max_t \left[ H_t \right], \quad \text{with} \quad H_t = -\sum_{i=1}^N p_{t,i} \log p_{t,i},
\end{equation}
where \(H_t\) is the entropy at time \(t\), and \(p_{t,i}\) are normalized prediction errors:
\begin{equation}
    p_{t,i} = \frac{\mathcal{L}_E(i)}{\sum_{j=1}^N \mathcal{L}_E(j)}, \quad \mathcal{L}_E(i) = \|\mathbf{f}_{t+1,i} - \hat{\mathbf{f}}_{t+1,i}\|^2.
\end{equation}
Peaks in the entropy curve \(H_t\) highlight moments of heightened prediction error variability, which are treated as event boundaries. 
For summarization, the most representative frame \(S_k\) for each segment \(k\) is selected by minimizing the prediction loss within the segment \([B_k, B_{k+1}]\):
\begin{equation}
    S_k = \arg \min_{t \in [B_k, B_{k+1}]} \mathcal{L}_E(t).
\end{equation}
Here, \(S_k\) corresponds to the frame with the lowest prediction error, representing the event's most stable and well-predicted observation. Empirically, in Table~\ref{tab:event_segmentation_sim}, we see that having a window of $30$, corresponding to the streaming video FPS, works best, and traditional peak detection directly on the prediction error leads to over-segmentation. 

\section{EXPERIMENTAL SETUP}
We evaluate the EASE framework in both simulated and real-world environments to assess its performance across active tracking, event segmentation, and summarization tasks.

\textbf{Simulation Environment.} Training and evaluation are conducted in UnrealCV-Gym. We train on the FlexibleRoom environment for real-world experiments, which offers adjustable clutter and difficulty settings. This environment provides dynamic lighting, cluttered scenes, and human-like actors performing diverse activities, enabling fine-grained control over agent actions and environmental dynamics. FlexibleRoom ensures a challenging yet controlled platform for training models that generalize effectively to real-world scenarios. We train on the Random Room environment and evaluate on the City1 and UrbanCity environments. 

\textbf{Real-world Evaluation.} Real-world experiments are performed using the Interbotix LoCoBot platform, equipped with a six-degree-of-freedom robotic arm and visual sensors. These experiments take place in an open space that simulates an office-like setting with distractors such as large windows, furniture, and stairs. Three actors perform unscripted actions typical of daily life—e.g., walking, adjusting thermostats, opening doors, tying shoelaces, and working on laptops—over 4-minute episodes comprising at least 10 actions, averaging 15 seconds each. Three annotators review video recordings to mark event boundaries, assess tracking success, and evaluate summarization quality, ensuring diverse and realistic conditions for robust evaluation. 

\textbf{Quantitative Evaluation Metrics.} Following prior work~\cite{shou2021generic}, event segmentation is evaluated using precision, recall, and F1 score, comparing detected boundaries to ground truth within tolerance windows. Strict evaluation uses narrow tolerances (e.g., 2 to 15 frames, corresponding to 0.5 seconds at 30 FPS), while relaxed evaluation allows broader tolerances (15 to 45 frames), reflecting the complexities of active event perception. Tracking performance in the simulation environment is measured using total environment reward and average episode length, following prior work~\cite{zhong2018advat,zhong2021advat}. For real-world tracking evaluation, we use the average qualitative judgment from the annotators, who grade each frame as 1 (tracking) or 0 (not tracking). 

% \clearpage
\begin{table}[t]
% \centering
\caption{Event Segmentation Results in Simulation Environments. }
\label{tab:event_segmentation_sim}
\resizebox{\columnwidth}{!}{
\begin{tabular}{|l|c|c|c|c|c|c|c|}
\hline
\multirow{2}{*}{\textbf{Model $\downarrow$ Env. $\rightarrow$}} & \textbf{Seg.}       & \multicolumn{2}{c|}{\textbf{City}} & \multicolumn{2}{c|}{\textbf{Urban City}} & \multicolumn{2}{c|}{\textbf{Rand. Room}} \\ \cline{3-8}
                     & \textbf{Mode} & \textbf{IoU} & \textbf{Acc} & \textbf{IoU} & \textbf{Acc} & \textbf{IoU} & \textbf{Acc} \\ \hline
EASE-Hybrid & 1 & 0.52 & 0.70 & 0.35 & 0.49 & 0.47 & 0.64 \\ 
EASE & 1  & 0.41 & 0.59 & 0.33 & 0.49 & 0.51 & 0.69 \\ 
\hline
\hline
EASE-Supervised  & 2   & 0.31 & 0.46 & 0.29 & 0.39 & 0.33 & 0.46 \\ 
EASE-Hybrid & 2 & 0.30 & 0.42 & 0.20 & 0.32 & 0.23 & 0.30 \\ 
EASE & 2 & 0.20 & 0.33 & 0.26 & 0.36 & 0.24 & 0.34 \\ 
\hline
\end{tabular}
}
*Segmentation modes (Seg. Mode) 1 and 2 denote how events are segmented. 1: using $\mathcal{L}_E$ or using prediction assessment.
\end{table}

\textbf{Baselines.} We evaluate three versions of our framework: (i) \textbf{EASE}, the fully self-supervised model, (ii) \textbf{EASE-Hybrid}, trained with both self-supervised losses and simulation rewards for enhanced tracking, and (iii) \textbf{EASE-Supervised}, trained solely on environmental rewards, representing state-of-the-art active tracking methods. All models share the same architecture. Segmentation and summarization for EASE and EASE-Hybrid use $\mathcal{L}_E$. For EASE-Supervised, we use state transition differences in the controller LSTM’s hidden state as the perception signal, as done in Predictability Assessment~\cite{shou2021generic}.

\section{RESULTS AND DISCUSSION}
% In this section, we present our results from the quantitative evaluation of the proposed EASE framework on both simulation (Section~\ref{sec:sim_results}) and real-world (Section~\ref{sec:real_results}) experiments. In Section~\ref{sec:qual}, we present a qualitative analysis of the framework, its strengths, and limitations. 

\subsection{Evaluation in Simulated Environments}\label{sec:sim_results}
To validate the framework, we first train and evaluate our model and baselines in diverse, progressively challenging environments within UnrealCV-Gym. Training is conducted in the Random Room environment, which provides a controlled setting with varying visual features such as illumination (color, direction, intensity) and textured backgrounds. As proposed in Luo \emph{et al.}~\cite{luo2019end}, random augmentation is applied using version v4 for training and v0 for testing. Additionally, the City1 and Urban environments, with increased clutter and distractions, are used to evaluate the model's robustness and adaptability.

\begin{table}[t]
\caption{Performance Evaluation on the Active Tracking Task.}
\label{tab:activeTracking}
\resizebox{\columnwidth}{!}{
\begin{tabular}{|l|c|c|c|c|c|c|}
\hline
\centering
\multirow{2}{*}{\textbf{Model $\downarrow$ Env. $\rightarrow$}} & \multicolumn{2}{c|}{\textbf{City}} & \multicolumn{2}{c|}{\textbf{Urban City}} & \multicolumn{2}{c|}{\textbf{Rand. Room}} \\ \cline{2-7}
               & \textbf{AR}     & \textbf{AL}    & \textbf{AR}     & \textbf{AL}    & \textbf{AR}      & \textbf{AL}    \\ \hline
TLD+PID~\cite{trehan2022towards} & 12 & \textbf{90} & 19 & \textbf{115} & \underline{25} & \textbf{147}\\
MIL+PID~\cite{trehan2022towards} & \textbf{32} & 59 & 24 & 50 & 21 & 43 \\
MOSSE+PID~\cite{trehan2022towards} & \underline{16} & \underline{56} & \textbf{49} & \underline{68} & \textbf{28} & \underline{62} \\
\hline
\hline
Smart-Target~\cite{zhong2018advat} & 232 & 473 & 233 & 466 & 403 & 458\\
Random-Target~\cite{zhong2018advat} & 214 & 455 & 204 & 464 & 409 & 455\\
AD-VAT+~\cite{zhong2021advat} & \textbf{326} & \underline{483} & \textbf{322} & \underline{488} & \underline{427} & \underline{493}\\
{EASE-Supervised}     & \underline{248}           & \textbf{500}            & \underline{290}           & \textbf{490}          & \textbf{459}            & \textbf{500}            \\ \hline\hline
PredLearn-PID~\cite{trehan2022towards} & 114 & 343 & 71 & 349 & 115 & 319 \\
\hline
{EASE-Hybrid} & \underline{233}           & \textbf{500}            & \textbf{253}           & \textbf{496}          & \textbf{438}            & \textbf{500}            \\ 
\hline
{EASE}  & \textbf{273}           & \textbf{500}            & \underline{155}           & \underline{443}          & \underline{214}            & \underline{491}          \\ \hline

\end{tabular}
}
\end{table}

\begin{table*}[t]
\centering
\caption{Real-World Performance Evaluation of EASE for Active Event Perception Tasks.}
\label{tab:real_world_segmentation}
\begin{tabular}{|c|c|c|c|c|c|c|c|c|c|c|}
\hline
\multirow{2}{*}{\textbf{Model}}       & \multicolumn{3}{c|}{\textbf{Segmentation (Strict)}} & \multicolumn{3}{c|}{\textbf{Segmentation (Relaxed)}} & \multicolumn{3}{c|}{\textbf{Summarization}} & \textbf{Tracking} \\ 
\cline{2-10}
                     & \textbf{Precision} & \textbf{Recall} & \textbf{F1}        & \textbf{Precision} & \textbf{Recall} & \textbf{F1}      
                     & \textbf{Coverage} & \textbf{Redundancy} & \textbf{Quality}
                     & \textbf{Success (\%)}     \\ \hline
\textbf{EASE}        & 14.71            & \textbf{47.62}         & \underline{22.47}            & 24.75            & \textbf{60.12}         & 35.07           & 
\textbf{4.58} & 3.58 & \underline{4.17}
& \underline{94.42}                   \\ 
\textbf{EASE-Hybrid} & \textbf{38.31}            & \underline{27.82}         & \textbf{32.06}            & \textbf{51.67}            & \underline{36.98}         & \textbf{42.88}       &     
4.08 & \textbf{4.17} & \textbf{4.28}
& 90.99                   \\ 
\textbf{EASE-Supervised} & 21.87        & 21.54         & 21.64            & \underline{37.04}            & 35.39         & \underline{36.05}          &  
\underline{4.27} & \underline{4.05} & 4.12
& \textbf{98.01}                   \\ \hline
\end{tabular}
\end{table*}

\subsubsection{Event Perception with Growing Amounts of Clutter}
We begin by evaluating EASE's ability to actively perceive events in controlled simulation environments by tracking people and segmenting their actions into constituent, atomic events. Specifically, we test the event perception capabilities in environments where actors perform simple action sets (e.g., walk, turn, stop) under varying levels of clutter and distractions. This assessment allows us to examine the model's adaptability to dynamic conditions and its proficiency in detecting simple behavioral transitions. These results lay the groundwork for demonstrating EASE's scalability and robustness and offer a controlled alternative to its evaluation in complex, real-world scenarios explored in Section~\ref{sec:real_results}. 

Table~\ref{tab:event_segmentation_sim} summarizes the performance of the baselines across three simulation environments - City, Urban City, and Random Room - quantified by IoU (segmentation accuracy) and frame-level accuracy (Acc), following prior works in event perception~\cite{Aakur2018APP}. IoU measures the overlap between predicted and ground-truth event segments, while Acc computes the mean frame-wise agreement between streaming ground-truth labels and predicted segmentation boundaries to measure temporal alignment. 
We can see that EASE-Hybrid achieves higher IoU and Acc scores in simpler environments (City), likely due to training that combines self-supervised and reward-driven training. 
However, as complexity increases (Urban City and Random Room), EASE demonstrates robust adaptability by maintaining competitive performance despite being entirely unsupervised. 
The results demonstrate the robustness and adaptability of the EASE framework, particularly the effectiveness of self-supervised learning (using $\mathcal{L}_E$) for event segmentation and active perception tasks. 

\subsubsection{Active Object Tracking}
In addition to assessing EASE's event perception capabilities, we explicitly evaluate its performance in active object tracking, a task that involves dynamically following salient actors while adapting to changing environments. Table~\ref{tab:activeTracking} summarizes the results, comparing EASE against state-of-the-art methods, including traditional trackers with PID controllers (top section), reinforcement learning-based approaches (middle section), and predictive learning frameworks (bottom section). 
We see that EASE outperforms traditional PID-based trackers and matches or exceeds the performance of existing reinforcement learning-based approaches, particularly in more challenging environments. While EASE-Hybrid benefits from reward-driven learning for simpler settings, the fully self-supervised EASE model demonstrates robust adaptability and consistent performance across diverse environments. 
This is remarkable, considering that EASE is not explicitly trained for tracking or event perception. %, and strikes a balance between perception and control.
% , providing a strong foundation for active event perception in real-world scenarios with more complex and diverse actions.

\subsection{Real-world Event Perception}\label{sec:real_results}
We extend the evaluation of EASE to real-world scenarios to assess its performance on active event segmentation and summarization tasks. Conducted in a dynamic, office-like environment with unscripted human activities, these studies aim to validate EASE's adaptability, self-supervised capabilities, and practicality in handling complex and diverse real-world actions. Table~\ref{tab:real_world_segmentation} summarizes the results across segmentation (strict and relaxed), summarization, and tracking success. 

\textit{Temporal Event Segmentation.}
Table~\ref{tab:real_world_segmentation} presents the segmentation results for EASE, EASE-Hybrid, and EASE-Supervised, evaluated under the strict and relaxed settings described earlier. The strict setting emphasizes precision by requiring accurate boundary predictions within narrow temporal tolerances, while the relaxed setting provides leeway, reflecting the complexity of real-world scenarios with diverse human activities and noisy motion patterns. 
The hybrid model achieves higher precision in both evaluation settings by predicting fewer boundaries and avoiding over-segmentation, though this comes at the cost of lower recall, particularly in segments with frequent transitions. In contrast, the fully self-supervised EASE model is highly sensitive to changes in both human and robot movements, resulting in more predictions and higher recall but lower precision due to over-segmentation. The supervised model balances these extremes with moderate precision and recall.%, predicting more boundaries than the hybrid model but fewer than the self-supervised one. 
\ While over-segmentation may initially appear as a limitation, it could prove beneficial for active event perception tasks. Studies from cognitive science~\cite{zacks2001event} suggests that human actions possess an inherently hierarchical structure, with events unfolding across multiple temporal scales. EASE’s fine-grained predictions may provide a foundation for encapsulating this hierarchy, presenting an avenue for future work to explore multilevel segmentation approaches.

\begin{figure*}
\centering
\setlength\tabcolsep{1pt}
\label{fig:qual}
    \begin{tabular}{cccccc}
    \hline
    \multicolumn{6}{c}{\textbf{Learning to move backwards when too close to the target}}\\
    \hline
         \includegraphics[width=0.16\textwidth]{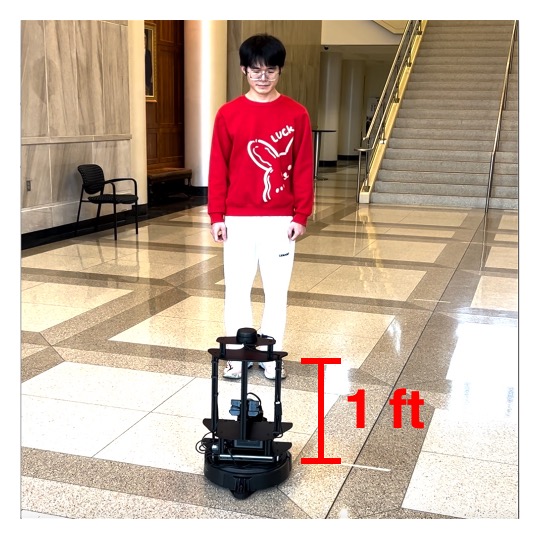} & 
         \includegraphics[width=0.16\textwidth]{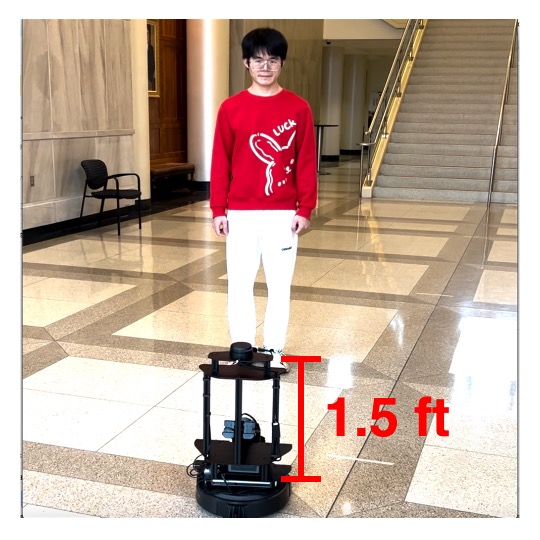} & 
         \includegraphics[width=0.16\textwidth]{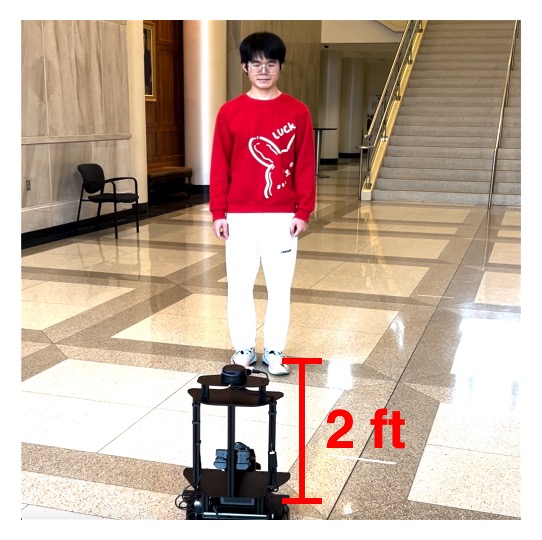} & 
         \includegraphics[width=0.16\textwidth]{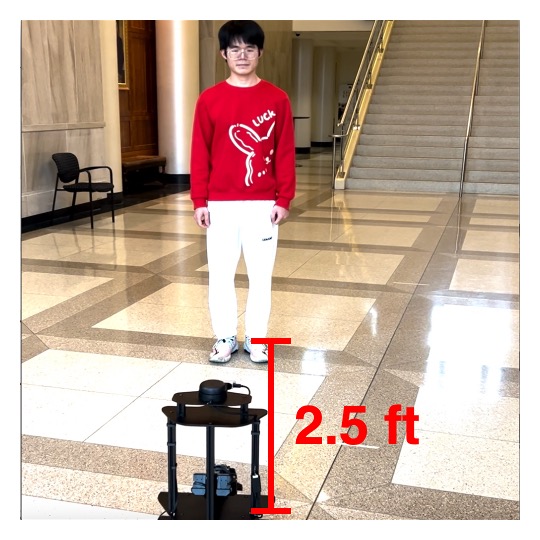} & 
         \includegraphics[width=0.16\textwidth]{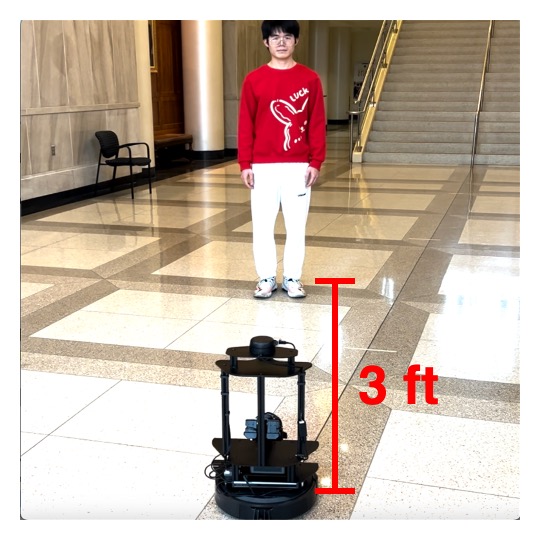} & 
         \includegraphics[width=0.16\textwidth]{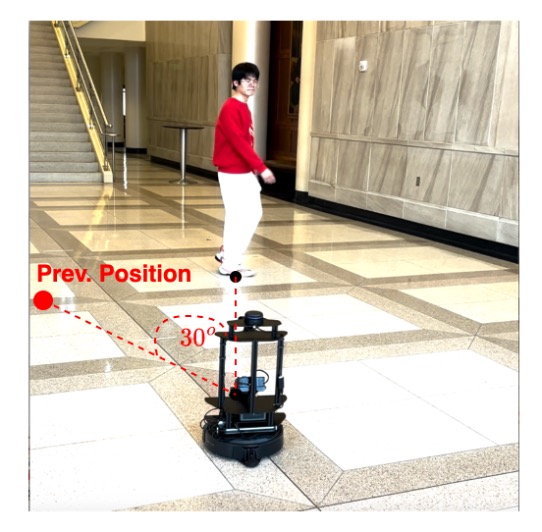} \\
         \hline
    \multicolumn{6}{c}{\textbf{Reacting to a new actor in addition to the target in the scene}}\\
    \hline
         \includegraphics[width=0.16\textwidth]{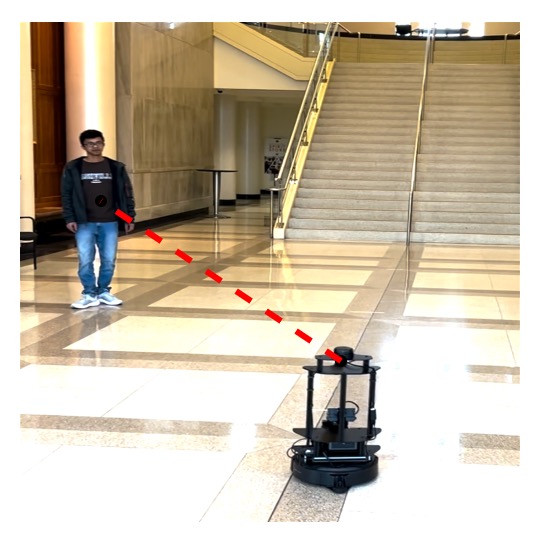} & 
         \includegraphics[width=0.16\textwidth]{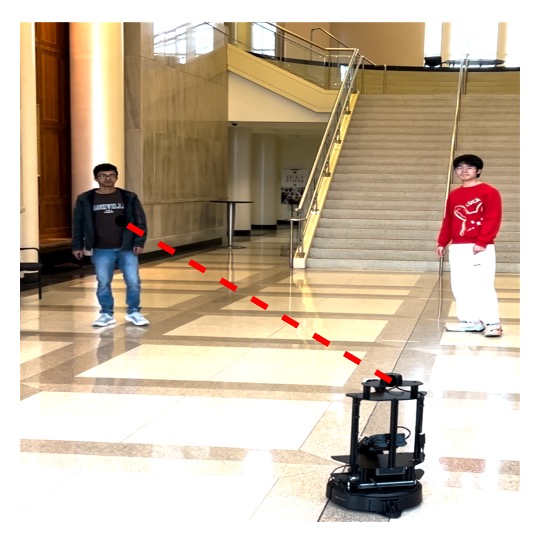} & 
         \includegraphics[width=0.16\textwidth]{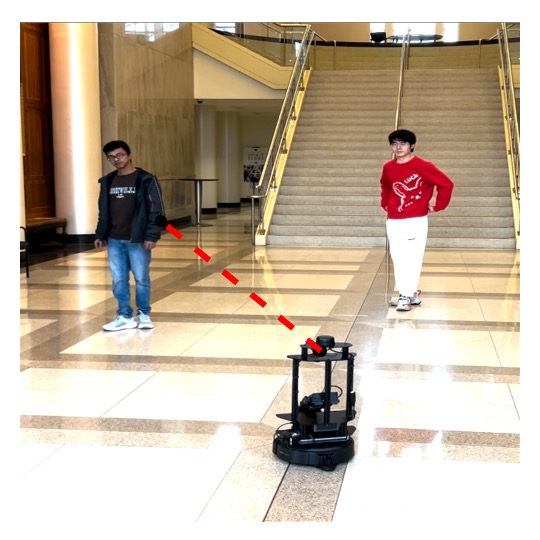} & 
         \includegraphics[width=0.16\textwidth]{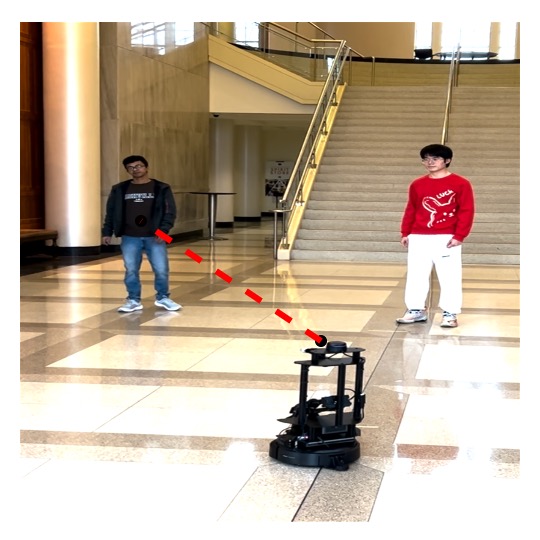} & 
         \includegraphics[width=0.16\textwidth]{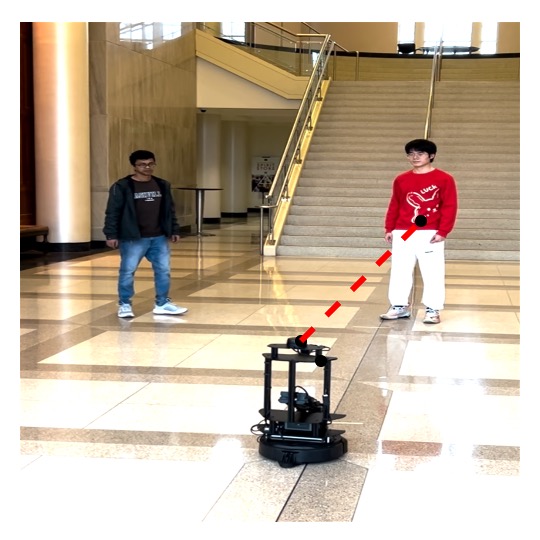} & 
         \includegraphics[width=0.16\textwidth]{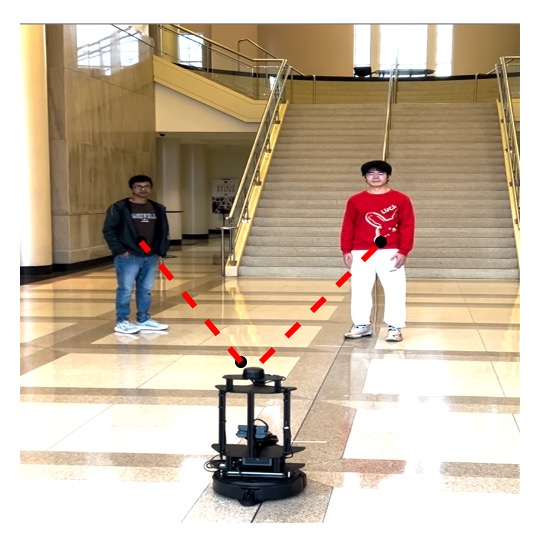} \\
        \hline
    \multicolumn{6}{c}{\textbf{Summarization results of action ``Move chair to the other side''}}\\
    \hline
    \multicolumn{6}{c}{\includegraphics[width=0.99\textwidth]{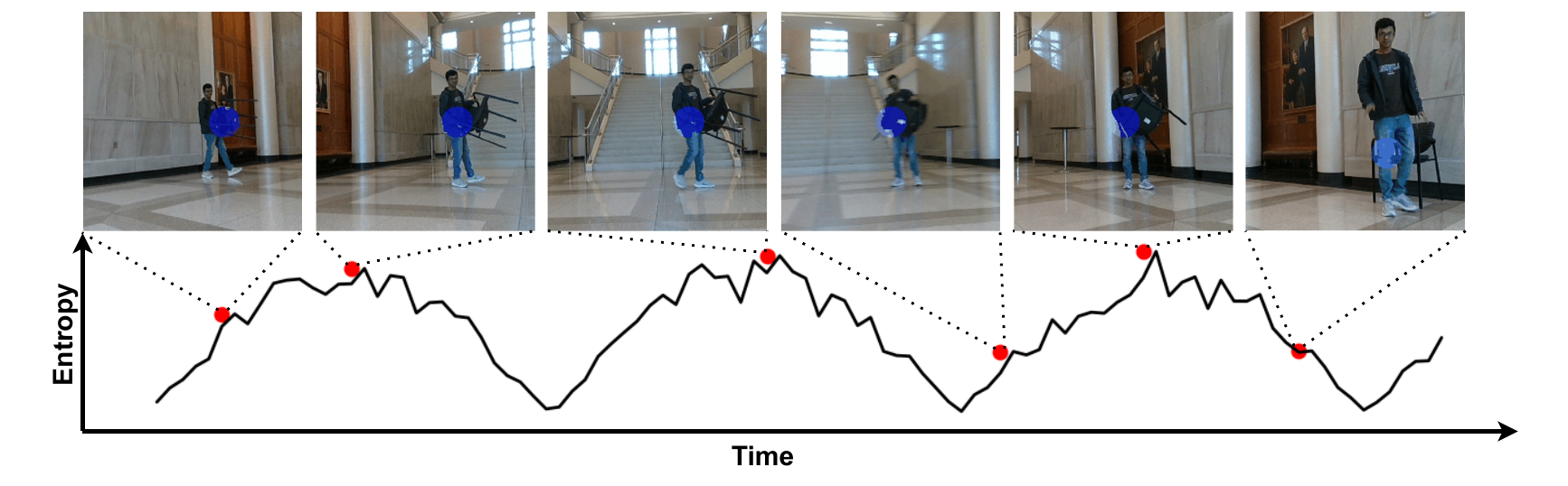}}
    \end{tabular}
    \caption{\textbf{Qualitative visualization} of emergent properties from free energy minimization. Top: EASE learns to move back when too close to the target. Middle: When a new actor is introduced, EASE keeps track of the old target until movement becomes predictable, reacts to the new target, and moves back to keep both actors in line of sight.
    Bottom: Entropy values and the summarization output (from robot POV) of the target performing action ``Move chair to other side'', where the over-segmentation of constituent actions (e.g., pick-up, walk, stop, and put-down) is visible. A blue circle indicates areas of high uncertainty. 
    }
\end{figure*}

\textit{Summarization}. 
Summarization complements event segmentation by distilling a continuous stream of segmented events into a concise and interpretable set of keyframes, enabling efficient review and analysis of embodied agent observations. Unlike conventional tracking methods that store raw video frames for post-processing, EASE performs summarization entirely in a streaming manner, preserving privacy by retaining only high-level event representations. This is particularly critical in applications such as assistive robotics and surveillance, where minimizing stored data reduces the risk of privacy breaches. By operating without persistent data storage, EASE eliminates the need for manual filtering or retrospective analysis while maintaining temporal context and relevance. 
To evaluate summarization performance, we consider three metrics based on human judgment: Temporal Coverage, which assesses how well the summary captures all significant events in the video; Redundancy, which evaluates the uniqueness of each selected keyframe; and Quality, which measures the clarity and relevance of the chosen keyframes to the observed events. Each metric is rated on a 1–5 scale, with higher scores indicating better summarization outcomes. 
As shown in Table~\ref{tab:real_world_segmentation}, EASE achieves the highest scores for Temporal Coverage, reflecting its ability to capture all significant events in the video. However, its sensitivity to motion changes occasionally results in slightly higher Redundancy compared to hybrid and supervised models. The supervised model excels in Quality, benefiting from structured learning, while the hybrid model balances coverage and redundancy effectively but sometimes overlooks finer details of events. Importantly, unlike other methods that require storing entire video sequences or extracted feature histories, EASE generates summaries without retaining raw data, reinforcing its potential as a privacy-preserving framework for real-world deployment.

\textit{Tracking}.  
Real-world tracking success (Table~\ref{tab:real_world_segmentation}) is measured using a subjective metric: human annotators assign a binary score (1 if the target is tracked, 0 otherwise) to each frame, with the mean score reported across the sequence. EASE achieves a 94.42\% success rate, reflecting robust continuity in tracking salient actors, while EASE-Supervised scores slightly higher (98.01\%). Along with the quantitative evaluation from  Table~\ref{tab:activeTracking}, these results indicate that EASE balances tracking fidelity and localization.

\subsection{Qualitative Analysis}\label{sec:qual}
EASE demonstrates notable emergent behaviors as a result of its free energy minimization objective, achieved without explicit action supervision or external rewards. 
Some examples are shown in Figure~\ref{fig:qual}. 
For instance, when the human target moves too close, EASE learns to take backward steps to re-establish the target within its field of view (FOV). When tracking is lost, it autonomously searches in the direction the target was last observed, exhibiting an implicit memory-like behavior. In contrast, the hybrid and supervised models quickly switch focus to new distractors, such as another person entering the frame, while the SSL model maintains continuity by sticking to the initial target. This sustained focus underscores the SSL model’s capacity for recognizing event continuity, albeit at the cost of over-segmenting repetitive motions. 
Figure~\ref{fig:qual} illustrates these emergent behaviors\footnote{See \href{https://saakur.github.io/Projects/EASE/index.html}{project page} for additional examples.} These behaviors highlight EASE's ability to adaptively couple perception and action to maintain focus on salient targets. However, the prediction-driven mechanism has its drawbacks. In scenarios with subtle or limited target motion, EASE occasionally prioritizes novel elements in the environment, such as stairs or furniture, over the primary target. 
% While this adaptability aids in cluttered scenes, it can lead to loss of focus on the original target.%, a trade-off not observed in the hybrid and supervised variants.

\section{CONCLUSIONS AND FUTURE WORK}
In this paper, we introduced EASE, a novel framework for active event perception that unifies spatiotemporal representation learning and embodied control through a free energy minimization paradigm. By leveraging self-supervised learning, EASE adaptively segments, summarizes, and tracks dynamic events in simulation and real-world environments without relying on annotations or extrinsic rewards. The quantitative and qualitative results highlight EASE's ability to balance fine-grained event sensitivity with robust motor control, giving rise to emergent behaviors such as implicit memory and continuity in tracking. 
% These findings underscore the potential of prediction-driven frameworks for achieving scalable and privacy-preserving event perception. 
In the future, we aim to enhance EASE by capturing the hierarchical nature of event segmentation, improving its adaptability to subtle or novel actions, and extending it to collaborative problem solving. 

% \textbf{Human Subjects Research.} This study did not involve external participants, and all data collection involved the authors themselves. No sensitive or private information was gathered, and no IRB approval was required.  

% \section{ACKNOWLEDGMENTS}
\textbf{Acknowledgements.} 
The authors would like to thank Mr. Joe Lin and Shanmukha Vellamcheti for helping with the hardware setup and feedback. 
ChatGPT and Grammarly were used to proofread and fix grammar errors. 
% \clearpage

\bibliographystyle{IEEEtran}
\bibliography{egbib}

\end{document}